\documentclass[runningheads]{llncs}
\usepackage[T1]{fontenc}
\usepackage{graphicx,verbatim}

\usepackage{amsmath,amsfonts,amssymb}
\usepackage{url}
\usepackage{multirow}
\usepackage{float}
\newcommand{\method}{MRSeg}
\newcommand{\R}{\mathbb{R}}
\newcommand{\GAP}{\operatorname{GAP}}
\newcommand{\meanpool}{\operatorname{Mean}}

\newcommand{\mean}{\operatorname{Mean}}
\newcommand{\topknorm}{\operatorname{TopKNorm}}
\newcommand{\reshape}{\operatorname{reshape}}
\newcommand{\softmax}{\operatorname{softmax}}
\begin{document}

\title{Multimodal Routing and Region Refinement for Language-Guided Medical Image Segmentation}
\titlerunning{MRSeg: Multimodal Routing and Region Refinement}
\author{
\mbox{Md Maklachur Rahman}\inst{1} \and
\mbox{Md Hasan Al Banna}\inst{1} \and
\mbox{Saraf Anjum}\inst{2} \and
\mbox{Assame Arnob}\inst{1} \and
\mbox{Tracy Hammond}\inst{1}
}

\authorrunning{M. M. Rahman et al.}

\institute{
Texas A\&M University, College Station, TX 77843, USA
\and
Independent Researcher
\\
\email{\{maklachur,mdhasanalbanna,assamearnob,hammond\}@tamu.edu},\\
\email{sarafanjumeva@gmail.com}
}

% index{Rahman, Md Maklachur}
% index{Hammond, Tracy}

% \author{Anonymized Authors}  %% Added for anonymized MICCAI submission
% \authorrunning{Anonymized Author et al.}
% \institute{Anonymized Affiliations \\
%     \email{email@anonymized.com}}
  
\maketitle              % typeset the header of the contribution

\begin{abstract}
Textual descriptions can reduce ambiguity in medical image segmentation by specifying the finding and location to be delineated. Existing text-guided methods mainly improve where image and language features interact but generally retain a single learned update pathway across all image--text pairs. We propose MRSeg, a parameter-efficient framework that uses each image–text pair to route the adaptation of visual and textual features before dense prediction. Frozen ConvNeXt-Tiny and PubMedBERT encoders provide multiscale visual features and clinical text tokens. A joint router uses the deepest visual feature and pooled text to predict a sparse mixture over low-rank adapter bases. The resulting route is shared across separate adapter banks for two visual scales and text, coordinating their adaptation while keeping the feature-specific parameters separate. Region Bridge uses text-derived queries to aggregate dense visual tokens into latent regions, refines these regions through self-attention and text cross-attention, and redistributes the refined information back to the feature maps. Finally, a multiscale decoder combines refined semantic features with shallow image evidence. On QaTa-COV19 and MosMedData+, MRSeg achieves 90.90/83.32 and 81.53/68.82 Dice/mIoU, respectively, with 7.11M trainable parameters and 7.60 GFLOPs.
Code: https://github.com/maklachur/MRSeg.

% Authors must provide keywords and are not allowed to remove this Keyword section.

\end{abstract}

\section{Introduction}
Medical image segmentation supports diagnosis, treatment planning, and longitudinal assessment by delineating anatomical structures and pathological regions. Encoder--decoder models such as U-Net~\cite{unet}, U-Net++~\cite{unet++}, and nnU-Net~\cite{nnunet} remain strong foundations, while transformer-based designs improve long-range context modeling~\cite{transunet,swinunet}, and recent state-space models such as Mamba provide an efficient alternative for capturing long-range dependencies~\cite{mambainvision}. Yet image-only segmentation~\cite{aulunet,mambaliteunet} remains difficult when abnormalities are diffuse, low contrast, heterogeneous in appearance, or visually similar to surrounding tissue. In these cases, the image may contain several plausible regions without explicitly indicating which finding is clinically relevant.

Clinical descriptions provide complementary information about the target, including its presence, laterality, location, number, or extent.
This motivates text-guided medical image segmentation, where an image and its associated description jointly determine the mask. LViT~\cite{lvit} established the foundation of language--vision interaction on pulmonary infection benchmarks. Later work explored encoder matching~\cite{mmiunet}, bidirectional and language-guided adapters~\cite{madapter,lga}, decoder attention~\cite{tgcam}, and reconstruction-based alignment~\cite{reclmis}. Related methods also use text-guided attention~\cite{tganet}, specialized decoders~\cite{ariadne}, memory and mixture-of-experts~\cite{mgunet,textmoe}, and robust prompting~\cite{arseg}. 
DD-CMD~\cite{ddcmd} introduced dual-domain cross-modal decoding that combines text-guided spatial attention with frequency-domain feature modulation.
These approaches demonstrate that text can sharpen target specification, but they primarily ask where and how the modalities should interact.

We therefore ask whether representation adaptation should be conditioned on the specific image--text pair. Although cross-attention is input-dependent, its learned update subspace remains shared across samples. Different cases may require distinct adaptation patterns. A small unilateral opacity may demand greater sensitivity to localized texture and position, whereas widespread infection requires broader spatial context. Fully fine-tuning both encoders, however, is computationally costly and prone to overfitting on limited medical vision--language data.

Another challenge is the granularity mismatch between language and dense segmentation. Clinical descriptions specify findings and anatomical locations at a semantic or regional level, whereas the output must delineate them at the pixel level. Direct token-to-pixel fusion asks a single interaction to identify the target, organize spatial evidence, and preserve boundaries. We instead decouple these roles. The image--text pair first determines how its representations should adapt; the adapted visual features are then organized and refined at the region level, and the decoder reconstructs the dense mask.

To address this, we propose MRSeg, a parameter-efficient framework for routed multimodal adaptation and region refinement. MRSeg retains frozen ConvNeXt-Tiny~\cite{liu2022convnext} and PubMedBERT~\cite{pubmedbert} backbones and learns lightweight LoRA~\cite{lora}, normalization, routing, refinement, and decoding parameters. A router derives a sparse policy from the globally pooled Stage-4 feature and mean-pooled text. Pair Adapter reuses this policy across separate low-rank banks for the Stage-3 feature, Stage-4 feature, and text tokens. The route is shared, but adapter parameters remain feature-specific. Two parallel Region Bridge modules form text-aware latent regions and return their refined information to the dense Stage-3 ($f_3$) and Stage-4 ($f_4$) maps through identity-safe residual updates.

% In summary, our contributions are threefold:
Our main contributions are summarized as follows:
(1) We propose MRSeg, a parameter-efficient framework that separates pair-conditioned adaptation, region-level refinement, and dense mask reconstruction using frozen visual and language backbones. (2) We introduce Pair Adapter, which applies a shared sparse image--text route across separate visual and textual low-rank banks, and Region Bridge, which refines dense features through text-aware latent regions. (3) Extensive experiments on QaTa-COV19 and MosMedData+ demonstrate the complementary benefits of both modules, achieving 90.90\%/83.32\% and 81.53\%/68.82\% Dice/mIoU with 7.11M trainable parameters and 7.60 GFLOPs.

\begin{figure}[t]
    \centering
    \includegraphics[width=0.95\linewidth]{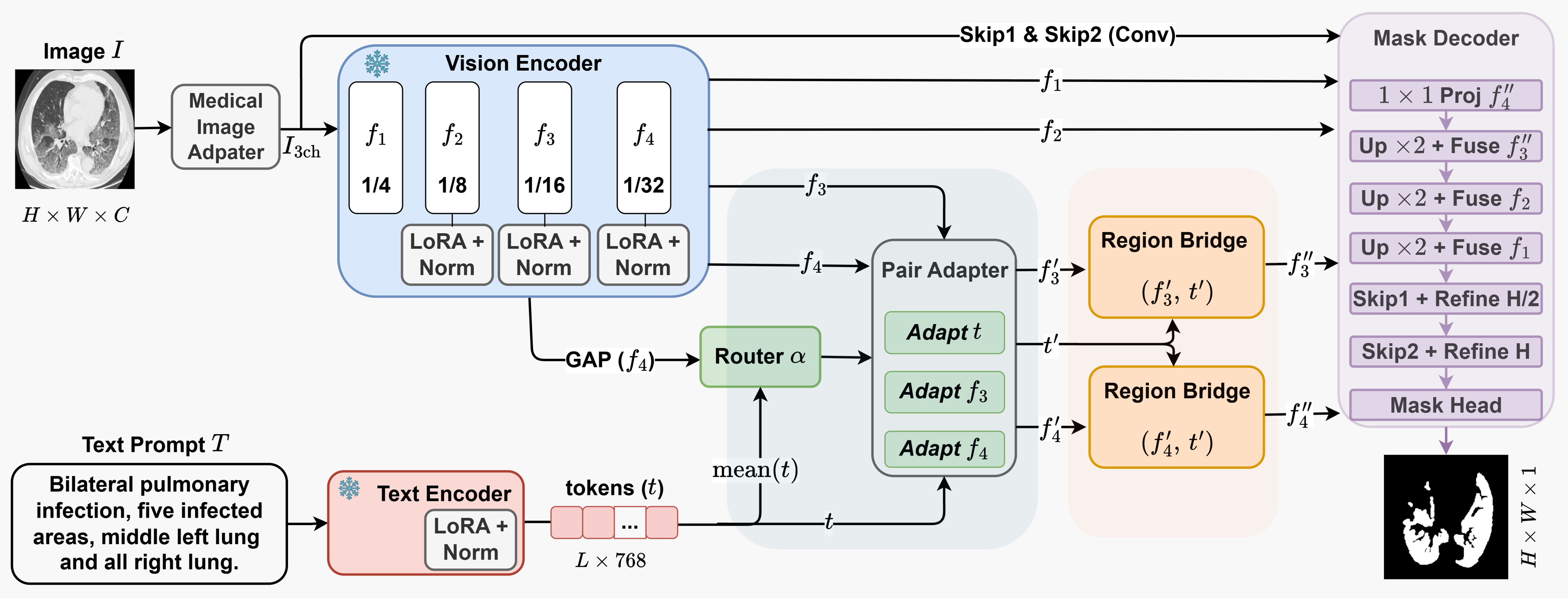}
    \caption{Overview of \method. Frozen, parameter-efficiently adapted encoders produce multiscale visual features and text tokens. The router uses only $\GAP(f_4)$ and $\meanpool(t)$ to derive a sparse route $\alpha$. Pair Adapter reuses this route across separate $f_3$, $f_4$, and text adapter banks. Two Region Bridge modules independently produce $f_3''$ and $f_4''$, which are decoded together with $f_1$, $f_2$, and shallow features from the adapted image.}
    \label{fig:pipeline}
\end{figure}

\section{Methodology}

The overall pipeline is illustrated in Fig.~\ref{fig:pipeline}, and Fig.~\ref{fig:components} details the two proposed components. The design follows three principles. First, pretrained encoders should retain their general visual and biomedical priors, so we freeze their base parameters and learn only lightweight residual updates. Second, the current image--text pair should determine which update subspaces are active. Third, cross-modal refinement should pass through an intermediate regional representation before dense decoding. 
Given an image $I\in\R^{H\times W\times C}$ and clinical text $T$, MRSeg predicts logits $\hat{Y}\in\R^{H\times W\times 1}$.
The complete forward path of our proposed method is the following:
\begin{equation}
\begin{aligned}
% I_3&=\mathcal{A}_{\rm img}(I),\\
% I_3 &= I_{\mathrm{base}} + \mathcal{A}_{\rm img}(I_{\mathrm{gray}}),\\
I_3 &= \mathcal{A}_{\rm adapt}(I),\\
(f_1,f_2,f_3,f_4)&=E_v(I_3),\qquad t=E_t(T),\\
\alpha&=\mathcal{R}\!\left(\GAP(f_4),\meanpool(t)\right),\\
(f_3',f_4',t')&=\mathcal{A}_{\rm pair}(f_3,f_4,t;\alpha),\\
f_s''&=\mathcal{B}_s(f_s',t'),\quad s\in\{3,4\},\\
\hat Y&=\mathcal{D}(I_3,f_1,f_2,f_3'',f_4'').
\end{aligned}
\label{eq:overview}
\end{equation}
The low-level features $f_1$ and $f_2$ retain local structure and skip the proposed adaptation modules. In contrast, the more semantic features $f_3$ and $f_4$ are processed by routed adaptation and region refinement. The two Region Bridge modules operate independently and in parallel.

\subsection{Vision and Language Encoding}
\subsubsection{Medical image adaptation:}
For grayscale inputs, direct channel replication satisfies the three-channel requirement of a natural-image backbone but cannot correct modality-specific intensity statistics. We use a residual Medical Image Adapter:
\begin{equation}
I_3=I_{\rm base}+\mathcal A_{\rm img}(I_{\rm gray}),
\label{eq:image_adapter}
\end{equation}
where $I_{\rm base}$ is the replicated or retained three-channel image. The correction branch uses pointwise projection, depthwise $3\times3$ filtering, normalization, GELU, and a final pointwise projection. The last projection is zero-initialized, so the module initially reproduces the standard three-channel path and learns only the correction required by the task.

\subsubsection{Vision encoder:}
A frozen ConvNeXt-Tiny~\cite{liu2022convnext} extracts four feature maps $f_1$, $f_2$, $f_3$, and $f_4$ with channels $\{96,192,384,768\}$ and strides $\{4,8,16,32\}$. LoRA~\cite{lora} is added into selected pointwise transformations in the later stages, and the corresponding normalization parameters are trainable. In the visual branch, the reduced-rank representation is additionally processed by depthwise $3\times3$ spatial mixing. This keeps the pretrained projection fixed while allowing local task-specific corrections. Fig.~\ref{fig:pipeline} denotes these updates compactly as ``LoRA + Norm.''

\subsubsection{Language encoder:}
The prompt is padded or truncated to $L=24$ tokens and encoded by frozen PubMedBERT, yielding $t\in\R^{L\times768}$. LoRA is applied to the query and value projections of the final four transformer layers, together with their normalization parameters. The full sequence is retained for region-to-text interaction, while $\bar t=\meanpool(t)$ provides a compact routing descriptor. Encoder LoRA learns a dataset-level adjustment, whereas the routed feature adapters introduced next provide sample-specific updates.

\subsection{Routed Multimodal Adaptation}
A fixed adapter is task-specific but not case-specific. The conventional adapter applies the same residual transformation to every sample. Pair Adapter instead maintains $M$ low-rank bases and lets each image--text pair activate a sparse subset. We first compute $\bar f_4=\GAP(f_4)$ and $\bar t=\mean(t)$. The router produces:
\begin{equation}
\boldsymbol\alpha=\topknorm\!\left(\softmax\!\left(g_\theta[\bar f_4\Vert\bar t]\right)\right)\in\R^{M},
\label{eq:router}
\end{equation}
where $\topknorm$ retains and renormalizes the $K$ largest weights. We use $M=4$ available bases and $K=2$ active bases. The router excludes $f_3$ as an input because $f_4$ provides the most compact global context, while $f_3$ remains a higher-resolution feature to be adapted.

\begin{figure}[t]
    \centering
    \includegraphics[width=0.9\linewidth]{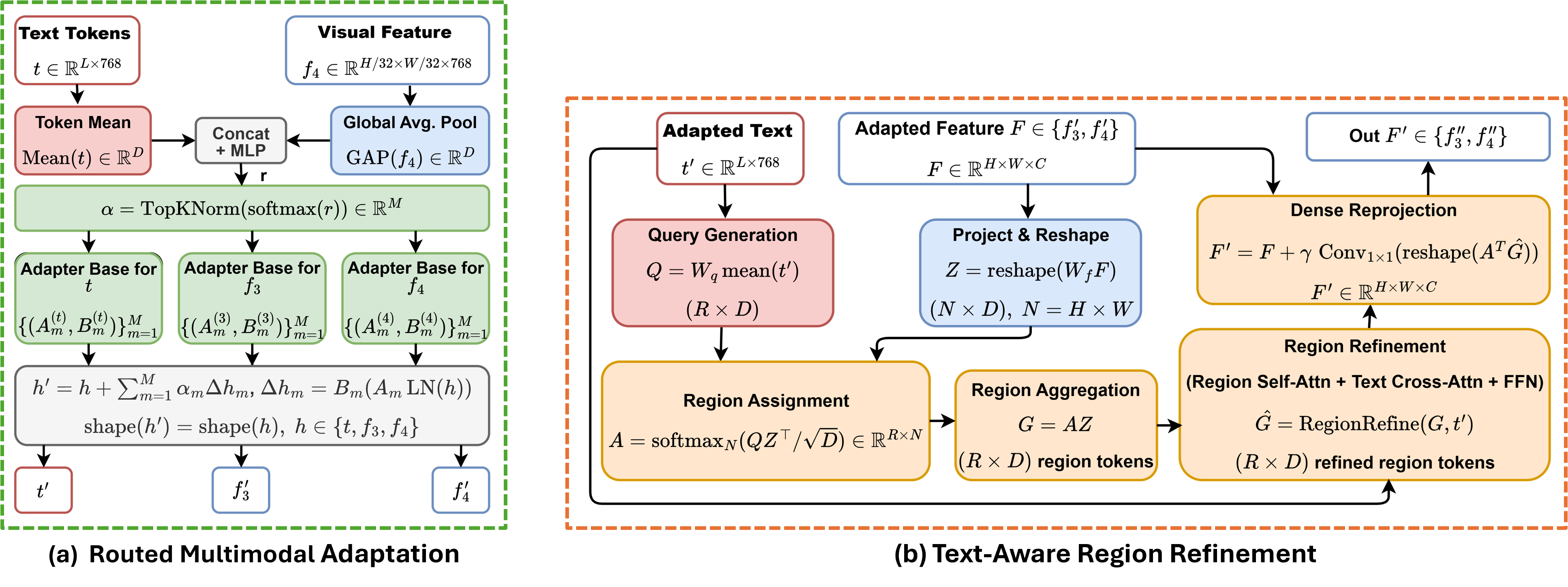}
    \caption{Core components of \method. (a) Routed multimodal adaptation: the image--text route controls separate adapter banks for $t$, $f_3$, and $f_4$. (b) Text-aware region refinement: text-derived queries aggregate dense visual tokens, regional and cross-modal interactions refine them, and dense reprojection returns the correction to the original feature map.}
    \label{fig:components}
\end{figure}

The same route $\alpha$ controls three separate adapter banks. For $h\in\{t,f_3,f_4\}$,
\begin{equation}
h'=h+\frac{\eta}{r}\sum_{m=1}^{M}\alpha_m B_m^{(h)}\!\left(A_m^{(h)}\operatorname{LN}(h)\right),
\label{eq:routed_adapter}
\end{equation}
where $r$ is the low-rank dimension and $\eta$ is the LoRA scale. Visual maps are flattened to token sequences and restored after adaptation. The key distinction is that the route is shared, but the adapter parameters are not. This encourages compatible specialization across modalities without forcing text and visual features through identical projections. All up-projections are zero-initialized, making Pair Adapter an identity mapping at initialization. Fig.~\ref{fig:components}(a) summarizes the route and its three applications.

\subsection{Text-Aware Region Refinement and Mask Decoding}
\subsubsection{Region formation and refinement:}
For scale $s\in\{3,4\}$, let $F=f_s'\in\R^{H_s\times W_s\times C_s}$ and $N=H_sW_s$. A $1\times1$ projection and reshaping produce $Z\in\R^{N\times D}$. Mean-pooled adapted text generates region queries $Q\in\R^{R_s\times D}$. We use $D=192$, $R_3=12$, and $R_4=6$. After LayerNorm, the assignment and aggregation are
\begin{equation}
A=\softmax_{N}\!\left(\frac{Q\,\bar Z^{\top}}{\sqrt D}\right)\in\R^{R_s\times N},
\qquad G=AZ,
\label{eq:region_assignment}
\end{equation}
where $\bar Z=\operatorname{LN}(Z)$. Softmax is applied over the $N$ spatial tokens, so each region query gathers a normalized distribution of visual evidence. The resulting region tokens are refined by two blocks, each containing region self-attention, cross-attention to projected $t'$, and a feed-forward network:
$\widehat G=\mathcal F_{\rm reg}(G,t')$.

The same assignment redistributes the refined tokens to the dense grid:
\begin{equation}
F'=F+\gamma P_{\rm out}\!\left(\reshape(A^{\top}\widehat G)\right),
\label{eq:dense_reprojection}
\end{equation}
where $\gamma$ is learnable. The output projection is zero-initialized, so Region Bridge initially preserves its input. Applying Eq.~\ref{eq:dense_reprojection} independently at the two scales gives $f_3''$ and $f_4''$. These latent regions are learned solely through the segmentation objective and should not be interpreted as supervised anatomical entities.

\subsubsection{Mask decoding and optimization:}
The decoder projects $f_4''$ to 192 channels, upsamples it, and fuses it with $f_3''$. It then introduces $f_2$ and $f_1$ through residual semantic skip fusion. Given decoder feature $x$ and skip $s$, a lightweight gate predicts $g=\tanh\mathcal{G}([\operatorname{Up}(x);s])$ and uses $\tilde s=s+s\odot g$. The gate output is zero-initialized, recovering ordinary skip fusion at the start of training. Two shallow stems extracted directly from $I_3$ provide half ($H/2$) and full-resolution ($H$) detail. A coarse $1\times1$ classifier and a depthwise detail branch produce additive logits, $\hat Y=\hat Y_{\rm coarse}+\Delta\hat Y_{\rm detail}$. We optimize equally weighted binary cross-entropy with logits and soft Dice loss: $\mathcal L=\mathcal L_{\rm BCE}+\mathcal L_{\rm Dice}$. Sigmoid and a $0.5$ threshold are used to produce the mask output.
\section{Experiments and Results}
\label{sec:experiments}
\subsubsection{Datasets, Evaluation Metrics, and Implementation Details:}
Following prior methods \cite{lvit,tgcam,reclmis,mgunet}, we evaluate on two public image--text--mask benchmarks for pulmonary infection segmentation. QaTa-COV19~\cite{covid19,lvit} contains 9,258 chest X-rays and is split into 5,716/1,429/2,113 as training, validation, and test samples. MosMedData+~\cite{mosmed,lvit} contains 2,729 CT slices and is split into 2,183/273/273 samples. We primarily report the Dice coefficient and mean Intersection over Union (mIoU) as overlap-based evaluation, and the 95th percentile of the Hausdorff Distance (HD95), measured in pixels, as the boundary error metric in the ablation studies.

All images are resized to $224\times224$, and descriptions use $L=24$ tokens. Training applies different augmentations, followed by channel-wise normalization. We use AdamW~\cite{adamw}, batch = 8, initial learning rate $10^{-4}$, weight decay $10^{-4}$, cosine annealing to $10^{-6}$, and early stopping with patience 60. QaTa-COV19 is trained for at most 200 epochs and MosMedData+ for 150. The LoRA rank is 8 with scale 16 and dropout 0.05. Both Region Bridges use two refinement blocks and four attention heads. Experiments run on one NVIDIA RTX 3090Ti (24GB).

\subsubsection{Comparison with State-of-the-Art Methods:}
\begin{figure}[t]
    \centering
    \includegraphics[width=\linewidth]{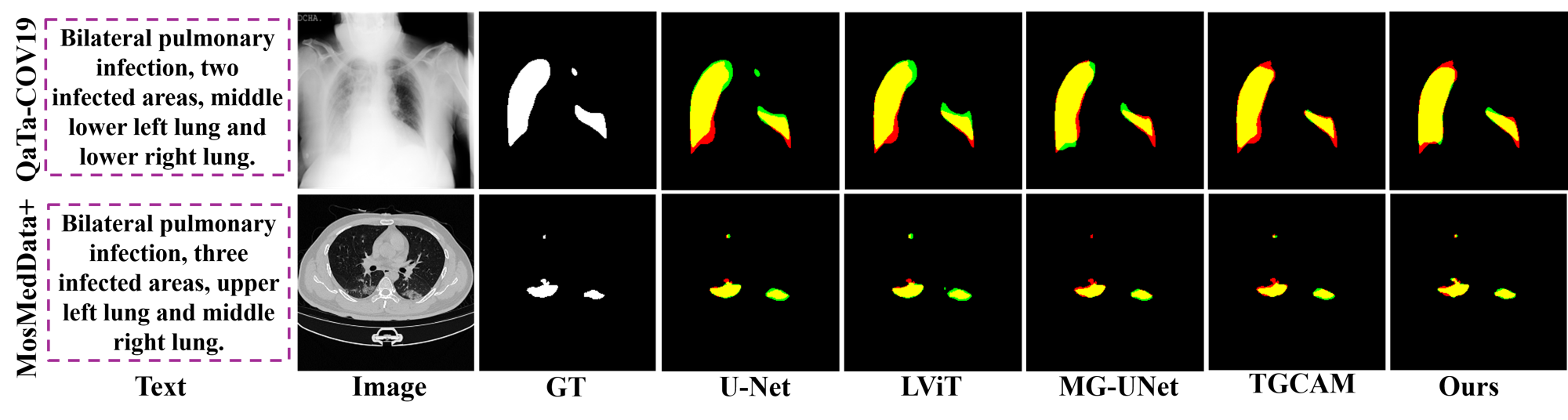}
\caption{Qualitative comparison on QaTa-COV19 and MosMedData+. Overlays indicate true positives (yellow), false negatives (red), and false positives (green).}
    \label{fig:qc_miccai26}
\end{figure}

Table~\ref{tab:main_comp} compares MRSeg with representative image-only and text-guided segmentation methods. MRSeg achieves the best performance on both datasets, obtaining 90.90\% Dice and 83.32\% mIoU on QaTa-COV19, and 81.53\% Dice and 68.82\% mIoU on MosMedData+. It improves over TGCAM~\cite{tgcam}, the strongest prior method on QaTa-COV19, by 0.30 Dice and 0.51 mIoU points. On MosMedData+, MRSeg surpasses the best prior Dice from MAdapter~\cite{madapter} by 2.91 points and the best prior mIoU from RecLMIS~\cite{reclmis} by 3.75 points. Importantly, MRSeg requires only 7.11M trainable parameters and 7.60 GFLOPs, demonstrating strong performance with parameter-efficient adaptation. The qualitative comparisons in Fig.~\ref{fig:qc_miccai26} further show more complete lesion coverage and fewer missed or spurious regions across both datasets.

\begin{table*}[t]
\centering
\footnotesize
\renewcommand{\arraystretch}{1.08}
\setlength{\tabcolsep}{4.2pt}
\caption{Comparison with SOTA methods on QaTa-COV19 and MosMedData+. Best and second-best results are shown in \textbf{bold} and \underline{underlined}, respectively. $\uparrow$/$\downarrow$ indicate higher/lower is better. — indicates values not reported by the original paper. }
\label{tab:main_comp}
\resizebox{\textwidth}{!}{
\begin{tabular}{l|c|c|cc|cc|cc}
\hline
\multirow{2}{*}{Method} &
\multirow{2}{*}{Venue} &
\multirow{2}{*}{Text} &
Params $\downarrow$ &
FLOPs $\downarrow$ &
\multicolumn{2}{c|}{QaTa-COV19} &
\multicolumn{2}{c}{MosMedData+} \\
\cline{6-7}\cline{8-9}
& & & (M) & (G) &
Dice $\uparrow$ & mIoU $\uparrow$ &
Dice $\uparrow$ & mIoU $\uparrow$ \\
\hline
U-Net~\cite{unet}
& MICCAI'15 & $\times$
& 14.8 & 50.3
& 79.02 & 69.46
& 64.60 & 50.73 \\

nnUNet~\cite{nnunet}
& Nature'21 & $\times$
& 19.1 & 412.7
& 80.42 & 70.81
& 72.59 & 60.36 \\

Swin-UNet~\cite{swinunet}
& ECCV'22 & $\times$
& 82.3 & 67.3
& 78.07 & 68.34
& 63.29 & 50.19 \\

\hline\hline

LAVT~\cite{lavt}
& CVPR'22 & $\checkmark$
& 118.6 & 83.8
& 79.28 & 69.89
& 73.29 & 60.41 \\

LViT~\cite{lvit}
& IEEE TMI'23 & $\checkmark$
& 29.7 & 54.1
& 83.66 & 75.11
& 74.57 & 61.33 \\

LGA~\cite{lga}
& MICCAI'24 & $\checkmark$
& \underline{8.24} & 381.1
& 84.65 & 76.23
& 75.63 & 62.52 \\

TGCAM~\cite{tgcam}
& MICCAI'24 & $\checkmark$
& — & —
& \underline{90.60} & \underline{82.81}
& 77.82 & 63.69 \\

MAdapter~\cite{madapter}
& MICCAI'24 & $\checkmark$
& — & —
& 90.22 & 82.16
& \underline{78.62} & 64.78 \\

RecLMIS~\cite{reclmis}
& IEEE TMI'24 & $\checkmark$
& 23.7 & 24.1
& 85.22 & 77.00
& 77.48 & \underline{65.07} \\

ARSeg~\cite{arseg}
& MICCAI'25 & $\checkmark$
& — & —
& 84.09 & 72.64
& 73.24 & 59.82 \\

TextMoE~\cite{textmoe}
& MICCAI'25 & $\checkmark$
& — & —
& 89.08 & 80.32
& 74.66 & 59.57 \\

MG-UNet~\cite{mgunet}
& MICCAI'25 & $\checkmark$
& 30.5 & \underline{11.0}
& 88.10 & 77.80
& 76.39 & 61.79 \\
\hline
\textbf{MRSeg (Ours)}
& -- & $\checkmark$
& \textbf{7.11} & \textbf{7.60}
& \textbf{90.90} & \textbf{83.32}
& \textbf{81.53} & \textbf{68.82} \\
\hline
\end{tabular}
}
\end{table*}

\subsubsection{Ablation Study on Different Core Modules:}
Table~\ref{tab:ablation} shows that each component contributes to MRSeg. Removing text causes the largest drop, confirming that clinical descriptions provide complementary guidance beyond the image alone. The routed adaptation and region bridges are also important, as removing either consistently degrades Dice, mIoU, and HD95 on both datasets. Independent image and text routers perform worse than our shared pair-conditioned router, supporting joint adaptation from the paired inputs. The medical image adapter and encoder LoRA provide further gains, while the full model achieves the best overall performance across all metrics.

\subsubsection{Ablation Study on Data Efficiency:}
As shown in Table~\ref{tab:data_efficiency}, our model remains competitive even under limited supervision. With only 30\% of the training data, MRSeg already approaches several state-of-the-art methods trained on the full dataset, achieving 89.60\% Dice on QaTa-COV19 and 77.83\% on MosMedData+. As we increase the training data, performance improves steadily and HD95 consistently decreases, showing that MRSeg learns effectively from limited annotations while continuing to benefit from additional supervision.

\begin{table}[t]
\centering
\footnotesize
\setlength{\tabcolsep}{3.6pt}
\renewcommand{\arraystretch}{1.05}
\caption{Ablation study on QaTa-COV19 and MosMedData+. Dice and mIoU are reported in \%, while HD95 in pixels. $w/o$ means without. Best values are bolded.}
\label{tab:ablation}
\resizebox{\linewidth}{!}{
\begin{tabular}{l|ccc|ccc}
\hline
\multirow{2}{*}{Model Variant} &
\multicolumn{3}{c|}{QaTa-COV19} &
\multicolumn{3}{c}{MosMedData+} \\
\cline{2-7}
& Dice $\uparrow$ & mIoU $\uparrow$ & HD95 $\downarrow$
& Dice $\uparrow$ & mIoU $\uparrow$ & HD95 $\downarrow$ \\
\hline
Image-only (No Text)
& 87.56 & 77.87 & 27.27
& 78.94 & 65.21 & 19.22 \\

$w/o$ Medical Image Adapter
& 90.68 & 82.95 & 16.45
& 80.99 & 68.06 & 15.18 \\

$w/o$ Encoder LoRA
& 90.49 & 82.64 & 16.90
& 80.76 & 67.73 & 16.11 \\

Independent Image/Text Routers
& 90.59 & 82.72 & 16.68
& 80.82 & 67.82 & 15.94 \\

$w/o$ Region Bridges
& 89.21 & 80.52 & 23.07
& 79.94 & 66.58 & 17.83 \\

$w/o$ Routed Adaptation
& 88.96 & 80.11 & 24.20
& 80.03 & 66.71 & 17.61 \\
\hline
Full model
& \textbf{90.90} & \textbf{83.32} & \textbf{14.34}
& \textbf{81.53} & \textbf{68.82} & \textbf{14.73} \\
\hline
\end{tabular}
}
\end{table}

\begin{table}[t]
\centering
\small
\renewcommand{\arraystretch}{1.1}
\setlength{\tabcolsep}{4pt}
\caption{Robustness under reduced supervision. Models are trained on randomly sampled \{30\%, 70\%, 100\%\} subsets of the training split and evaluated on the unchanged full test set. Dice and mIoU are reported in \%, and HD95 in pixels.}
\label{tab:data_efficiency}
\begin{tabular}{c|ccc|ccc}
\hline
\multirow{2}{*}{Training Data} &
\multicolumn{3}{c|}{QaTa-COV19} &
\multicolumn{3}{c}{MosMedData+} \\
\cline{2-7}
& Dice $\uparrow$ & mIoU $\uparrow$ & HD95 $\downarrow$
& Dice $\uparrow$ & mIoU $\uparrow$ & HD95 $\downarrow$ \\
\hline
30\%
& 89.60 & 81.17 & 17.07
& 77.83 & 63.71 &  18.53\\

70\%
& 90.48 & 82.62 & 15.57
& 80.11 & 66.83 & 17.94 \\

100\%
& \textbf{90.90} & \textbf{83.32} & \textbf{14.34}
& \textbf{81.53} & \textbf{68.82} & \textbf{14.73} \\
\hline
\end{tabular}
\end{table}

\section{Conclusion}
We presented MRSeg, a parameter-efficient framework that decomposes language-guided medical image segmentation into pair-conditioned adaptation, text-aware region refinement, and dense reconstruction. Pair Adapter coordinates feature-specific visual and textual updates through a shared sparse route, while Region Bridge organizes and refines visual evidence at the region level before mask decoding. Experiments and ablations on QaTa-COV19 and MosMedData+ demonstrate that these components provide complementary gains, enabling strong segmentation performance with only 7.11M trainable parameters and 7.60 GFLOPs. Building on these results, future work will extend evaluation beyond the current text-guided pulmonary infection benchmarks and examine robustness to variations in the input text.

% \begin{credits}
% % \subsubsection{\ackname} Thank you to the Sketch Recognition Lab (SRL) members for their continued support. This work was supported in part by NSF award 1948660.

% \subsubsection{\discintname}
% The authors have no competing interests to declare that are relevant to the content of this article.
% \end{credits}

\bibliographystyle{splncs04}
\bibliography{miccai26_bibliography}

\end{document}